\documentclass[pdflatex, sn-nature]{sn-jnl}

\usepackage{graphicx}%
\usepackage{multirow}%
\usepackage{amsmath,amssymb,amsfonts}%
\usepackage{amsthm}%
\DeclareMathAlphabet{\mathscr}{U}{rsfs}{m}{n}%
\usepackage[title]{appendix}%
\usepackage[table]{xcolor}%
\usepackage{textcomp}%
\usepackage{manyfoot}%
\usepackage{booktabs}%
\usepackage{algorithm}%
\usepackage{algorithmicx}%
\usepackage{algpseudocode}%
\usepackage{listings}%
\usepackage{float}

\usepackage{tabularx}
\usepackage{hyperref}
\usepackage[T1]{fontenc}
\usepackage[utf8]{inputenc}

\theoremstyle{thmstyleone}%
\theoremstyle{thmstyletwo}%

\theoremstyle{thmstylethree}%

\begin{document}

\title[Article Title]{Automated Extraction of Techno-Economic Data from 76,000 Energy System Studies}

\author*[1]{\fnm{Maxime} \sur{Gorres}}\email{m.gorres@fz-juelich.de}

\author[1]{\fnm{Jan} \sur{G\"opfert}}\email{j.goepfert@fz-juelich.de}

\author[1]{\fnm{Patrick} \sur{Kuckertz}}\email{p.kuckertz@fz-juelich.de}

\author[1]{\fnm{Noor Titan Putri} \sur{Hartono}}\email{t.hartono@fz-juelich.de}

\author[1,2]{\fnm{Heidi} \sur{Heinrichs}}\email{h.heinrichs@fz-juelich.de}

\author[1]{\fnm{Jochen} \sur{Linßen}}\email{j.linssen@fz-juelich.de}

\author[3]{\fnm{Iain} \sur{Staffell}}\email{i.staffell@imperial.ac.uk}

\author[1]{\fnm{Jann Michael} \sur{Weinand}}\email{j.weinand@fz-juelich.de}

\affil[1]{%
  \orgdiv{Institute of Climate and Energy Systems – J\"ulich Systems Analysis}, \orgname{Forschungszentrum J\"ulich GmbH},
  \orgaddress{\city{J\"ulich}, \postcode{52425}, \country{Germany}}}

\affil[2]{%
  \orgdiv{Department of Mechanical Engineering},
  \orgdiv{Chair of Energy Systems Analysis},
  \orgname{University of Siegen},
  \orgaddress{\city{Siegen}, \postcode{57076}, \country{Germany}}}

\affil[3]{%
  \orgdiv{Centre for Environmental Policy},
  \orgname{Imperial College London},
  \orgaddress{\city{London},  \country{UK}}}


\abstract{
Energy system models guide societally important decisions, but their credibility rests on quantitative assumptions that are difficult to source and audit. Meta-analyses can improve transparency and modeling practices, but the rapid growth of publications makes manual information extraction increasingly impractical. Consequently, databases are updated infrequently and efforts are often duplicated across research groups. Here, we demonstrate the highly accurate automated extraction of quantitative information from 76,000 energy system studies published since 2010. We compile 3.2~million structured quantitative data points together with 20~million associated metadata entries, spanning a broad spectrum of technologies, methodological approaches and system characteristics. Beyond providing input data for models, the resulting FAIR database make the energy systems literature itself analysable. We show where academic assumptions diverge from empirical observed data, and how research priorities vary at scale across technologies, regions and time. To facilitate broad use within the community, the database is provided through an interactive dashboard, enabling users to filter, analyse and download data according to their specific research needs.}




\maketitle

\section*{Introduction}\label{Sec-Introduction}
A successful energy system transformation utilizes decision support from energy system models. These computer models, however, require parameterization based on large-scale techno-economic data. False assumptions can lead to unintended bias in models and distort results~\cite{egli2019}. At the same time, modern prospective system analyses require increasingly large databases of technologies across multiple scenarios in order to assess uncertainties and possible future developments~\cite{frey2025,müller2026probabilisticidentificationtechnologytipping}. Relevant parameters include capital costs~\cite{schmidt2017future, sievert2024considering, wiser2021expert}, cost of capital~\cite{steffen2025global},  technical lifetimes~\cite{figgener2024multi}, efficiencies~\cite{green2025solar}, and learning rates~\cite{behrens2024reviewing}, which are often reported at different spatial and temporal scopes.

Identifying relevant studies and extracting suitable parameters is a time-consuming process, prone to inconsistency and incompleteness~\cite{li2024chapter, xu2022validity}. Furthermore, resulting data collections are difficult to reproduce and similar extraction efforts are frequently duplicated across -- or even within -- research groups~\cite{hatton2024global}. A comprehensive and structured collection of energy systems analysis data would reduce the effort required for model development and analysis, and substantially improve how techno-economic literature is synthesized and reused~\cite{wiese2019open}.

In recent years, several studies have addressed this challenge by compiling databases for subdomains of energy research. Examples include databases on perovskite solar cells ~\cite{jacobsson2022open}, battery materials~\cite{D2SC04322J}, and six electricity generation technologies~\cite{hatton2024global}. In these studies, data extraction was either performed manually, requiring substantial human efforts, or automated for a narrow field and application. For the perovskite database, for example, updates occur only irregularly and do not keep pace with the growing volume of publications, presumably due to the substantial time cost
required~\cite{nomad_perovskite2026}. Maintaining and continuously updating these databases remains a major challenge, as energy technologies evolve quickly and techno-economic data can become outdated within a short period of time. Extending such databases to the broader energy systems domain is even more challenging because of the rapidly growing volume of publications and the high heterogeneity of data in this field.

Recent advances in natural language processing, particularly large language models~(LLMs), have enabled automated, large-scale extraction of information from scientific literature. Related work in other domains includes the use of fine-tuned BERT models to automatically generate databases on Curie temperatures and band gaps ~\cite{gilligan2023rule}, battery materials~\cite{D2SC04322J}. More recently, LLMs have been applied to extract information from articles on $CO_2$ electrocatalytic reduction~\cite{chen2024large} and catalytic performance in $CO$ hydrogenation processes~\cite{SROR2026100741}. 

\begin{figure}[ht!]
    \centering
    \includegraphics[width=1\linewidth]{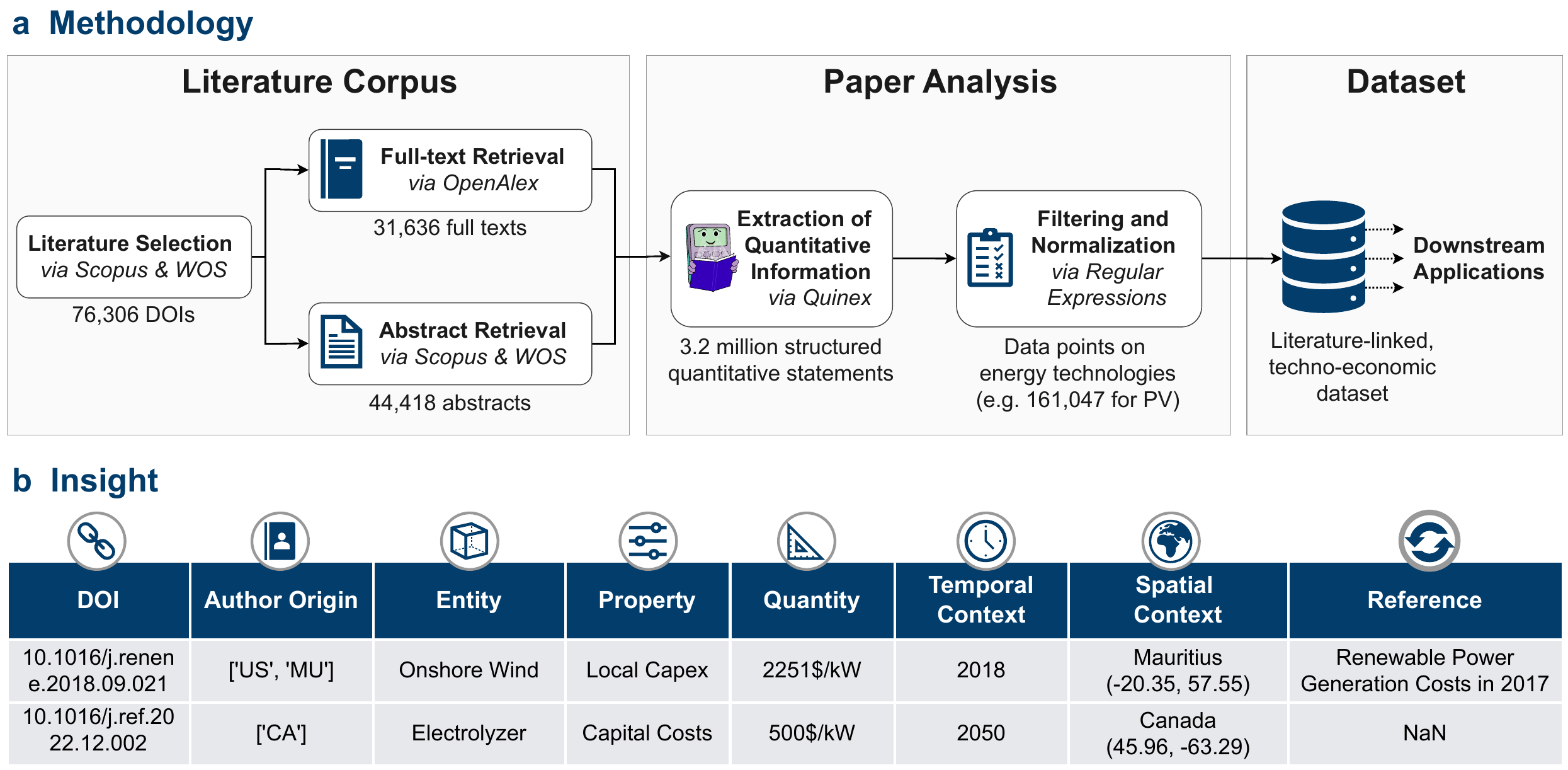}
    \caption{\textbf{Workflow for collecting the database from scientific literature.} \textbf{a} The workflow entails the collection of bibliographic metadata and text content (abstracts and full texts) from large-scale literature databases. Subsequently, quantitative information is automatically extracted using the domain-agnostic framework Quinex and normalized to predefined technologies and properties. \textbf{b} An insight into the collected metadata, such as the temporal context and spatial context, is given.}
    \label{fig:analysis-process}
\end{figure}

While previous studies have demonstrated the feasibility of LLM-based approaches for automated data collection, our work extends beyond these efforts by operating at a broader scope, without restricting the extraction to predefined entities or properties, while enriching the extracted structured information with the metadata required for energy systems analysis (see \hyperref[fig:analysis-process]{Fig.~1}). To this end, we employ Quinex~\cite{gopfert2026quinex}, a domain-agnostic framework for quantitative data extraction powered by lightweight LLMs. Quinex normalises the extracted quantities to ensure consistent units and improved database usability, while providing rich metadata that includes spatial and temporal information. By automating the extraction process, Quinex offers a scalable solution that facilitates continuous data updates and enables broad coverage across technologies, time periods and regions. 

To the best of our knowledge, this study presents the first large-scale database covering the full scope of energy systems analysis, with detailed metadata. We provide the resulting database to the community via an interactive dashboard, enabling easy access for analysis and reuse. In the following, we present an overview of the publication provenance and the technological composition of the database, then provide insights into specific temporal and geographical trends, demonstrating its coverage and consistency.


\section*{The first large-scale energy systems database}\label{Sec-Overview}

We extracted 3.2 million quantities, consisting of numerical values and units, along with 20 million associated metadata entries, including publication information such as DOIs, as well as the corresponding entities (e.g., solar PV) and properties (e.g. lifetime), from 76,000 publications in the energy systems analysis domain.

Beyond providing a source of quantitative energy systems data, the database also enables meta-analysis of the underlying scientific literature. Examining the geographical and technological composition of the corpus can reveal structural patterns in global research activity and thematic focus within the energy systems analysis community.

\begin{figure}[ht!]
    \centering

    \includegraphics[width=1\linewidth]{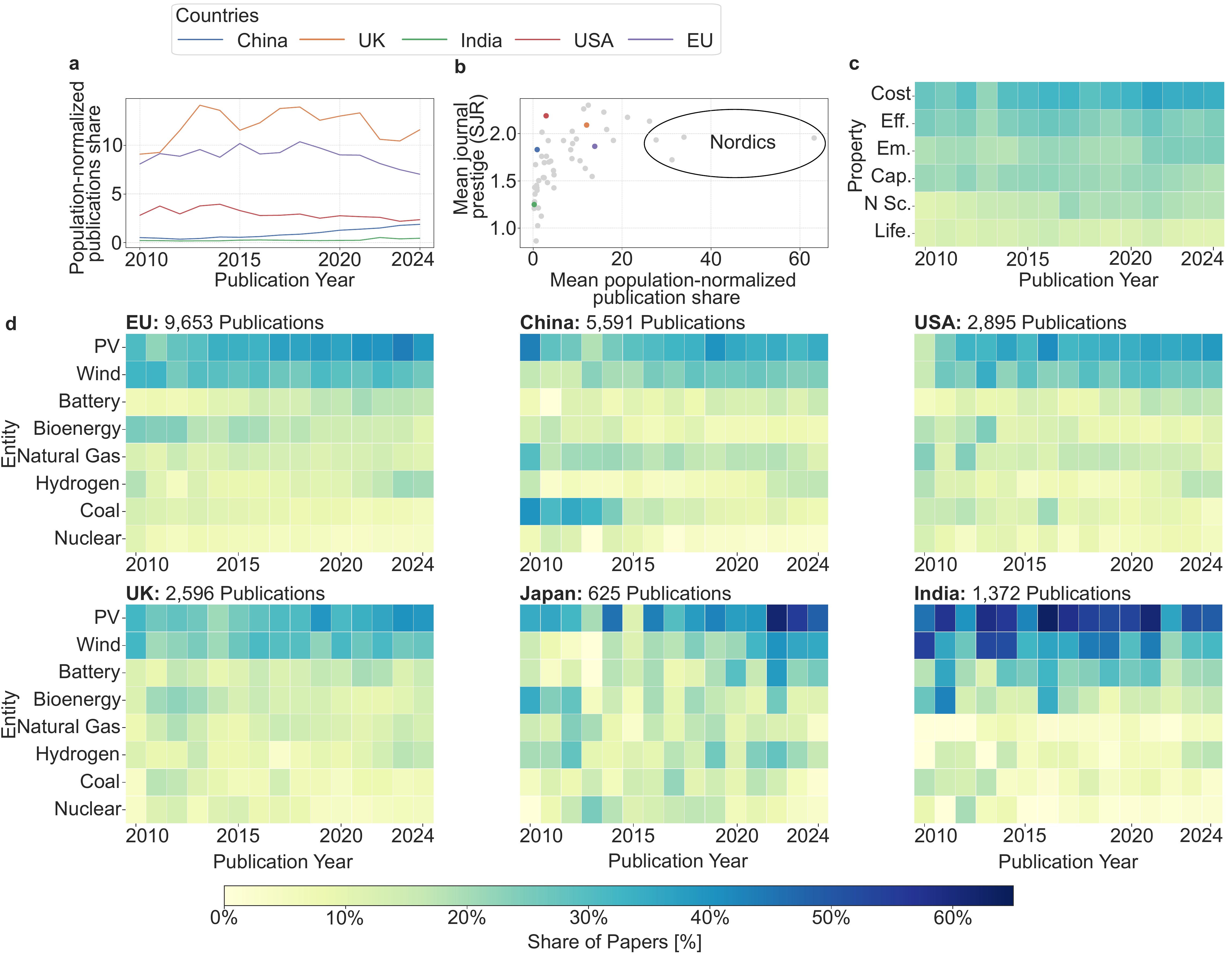}
    
    \caption{\textbf{Composition of the literature corpus of 25,545 papers with available country metadata.} \textbf{a} Temporal evolution of population-normalized publication share for the most represented countries and the European Union in its 2026 boundaries, calculated as the ratio between the annual share of publications divided by the corresponding share of the global population. Values above one indicate a publication output above the level expected from population size alone. \textbf{b} Relationship between population-normalized publication share and average journal visibility, measured as the publication-weighted mean SCImago Journal Rank (SJR)~\cite{scimago2025}. To approximate journal visibility at the country level, we assigned each publication the SJR of its journal and calculated the mean across all publications associated with that country. \textbf{c} Heat map illustrating the annual share of publications mentioning the six most frequently mapped properties across the literature corpus. \textbf{d} Heat maps showing the annual share of publications mentioning the mapped technologies across six geographical units (five countries and the European Union). All analyses in this figure are restricted to the retrieved full-text publications with available country metadata (n = 25,545). The affiliation country information was retrieved via the OpenAlex API~\cite{priem2022openalex}. Publications with multiple country affiliations were fully assigned to each contributing country and publications from 2025 were excluded because author affiliation information was unavailable for most countries. Country information was unavailable for around 6\% and SJR scores for around 4\% of studies. UK -- United Kingdom; USA -- United States of America; EU -- European Union; PV -- photovoltaics; Eff. -- Efficiency; Em. -- Emission; Cap. -- Installed Capacity; N. Sc. -- Number of Scenarios; Life. -- Lifetime.}
    \label{fig:literature-metadata}
\end{figure}

China contributes the largest absolute share of publications among individual countries, accounting for more than one in five studies (Supplementary Fig. S1); however, its population-normalized publication share is substantially lower (Fig.~\hyperref[fig:literature-metadata]{2a}). Notable shifts in publication activity can be observed over time. Since 2018, the population-normalized publication share has declined in Europe ($6.2\%/yr$) and, to a lesser extent, in the United Kingdom and the United States ($3.9\%/yr$ and $3.5\%/yr$, respectively), whereas it has continued to grow in China ($13.7\%/yr$). These trends indicate a gradual shift in the distribution of research activity within the field of energy systems analysis.

Several European countries, particularly the Nordic countries (Denmark, Norway, Sweden, Finland), exhibit high normalized publication shares (Fig.~\hyperref[fig:literature-metadata]{2b}), suggesting a strong energy systems analysis research concentration in the region. This reflects the strong national support for innovation in the renewable energy sector, for example through increased public energy R\&D funding, which has positioned the Nordic countries among the global leaders in this field~\cite{miremadi2019influence}. The United Kingdom produces more than ten times the number of publications expected based on its population size, while Denmark exceeds this benchmark by more than a factor of sixty.

Journal-level characteristics further highlight differences in publication patterns across countries (Fig.~\hyperref[fig:literature-metadata]{2b}). Some countries with a comparatively moderate publication share are represented in journals with high visibility, as measured by the mean SCImago Journal Rank (SJR)~\cite{scimago2025}. The Netherlands, for instance, exhibits the highest average visibility (SJR of 2.23) among countries contributing more than 2\% of publications. The United States of America combine a substantial share of the corpus (over 10\% of publications) with a high average visibility (2.19). 

Energy systems publications predominantly address photovoltaic and wind technologies, with a pronounced increase in studies on photovoltaics since 2014. Fossil-based technologies, including coal and natural gas, account for a comparatively smaller and decreasing share of publications throughout the observation period, whereas research on hydrogen has expanded in recent years. At the level of reported properties, cost- and efficiency-related indicators are most frequently documented (Fig.~\hyperref[fig:literature-metadata]{2c}).

Disaggregating the database by country reveals substantial temporal and thematic differences in research focus (Fig.~\hyperref[fig:literature-metadata]{2d}). In China, more than 30\% of publications between 2010 and 2013 included quantitative data on coal-based technologies, declining below 14\% since 2017. The gradual shift toward renewable energy technologies in subsequent years reflects the stronger renewable-energy deployment targets since the 12th Five-Year Plan~\cite{china_5year} adopted in 2011. The peak in PV-related research in 2010 may additionally be linked to the Golden Sun Programme~\cite{iea_golden_sun} introduced in 2009, which provided subsidies for solar power projects. European countries, by contrast, display a sustained emphasis on wind- and PV-related technologies throughout the observation period. In Japan, publications heavily focus on PV-related research, while research on wind technologies was limited in the early 2010s but increased after 2018. This trend is also reflected in actual deployment: as of April 2026, Japan had installed only 6~GW of wind capacity compared to 92 GW~of solar PV capacity~\cite{rei_japan_trends}. In contrast to European countries, Japan and India exhibit a stronger focus on battery related research. Furthermore, shifts in research focus can often be linked to major external events. For example, in Japan the share of publications on nuclear technologies peaks at 25\% in 2013, after the Fukushima nuclear accident.


\section*{Quantitative insights into model characteristics}\label{Sec-ModelInsights}

\begin{figure}[ht!]
    \centering
    \includegraphics[width=1\linewidth]{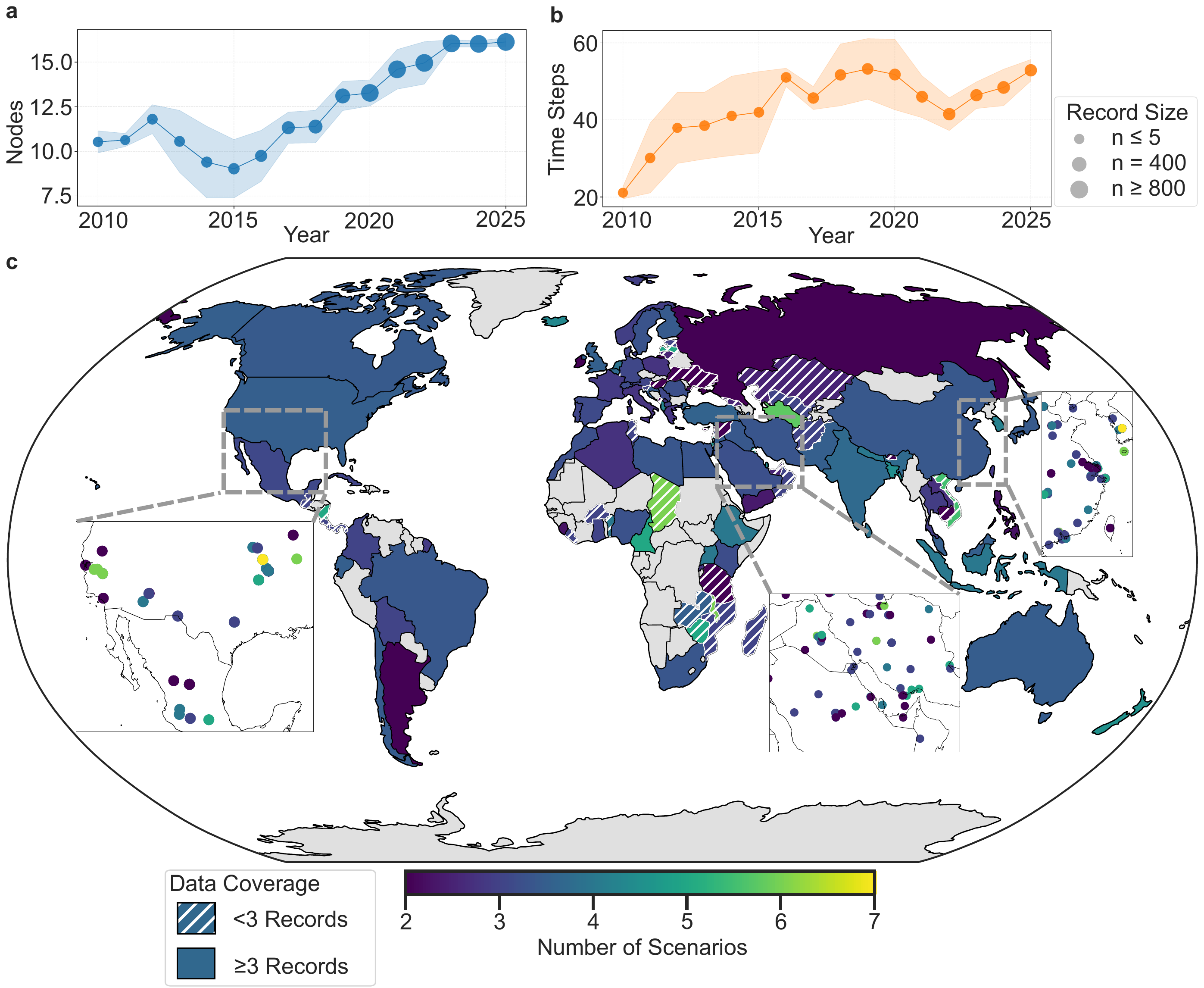}
    \caption{\textbf{Geographical and temporal distribution of the number of nodes, time steps and scenarios used in 4,970 energy systems modeling studies.}
    \textbf{a} and \textbf{b} show the number of nodes and the number of time steps reported in publications. \textbf{c} shows the country level average number of scenarios on a world map, while highlighting precise locations in Southwest Asia, East Asia and North America. Temporal values are shown as a centered 3-year moving average over the publication year. Only years with at least three observations from three distinct publications are retained. Shaded areas represent the standard error of the centered window, and point size reflects the number of observations per year. The geographic positions are derived from the spatial context extracted by Quinex. To reduce the influence of outliers, we applied Tukey's outlier detection method using a multiplier of 3 times the interquartile range (IQR)~\cite{tukey1977exploratory}.} 
    \label{fig:number_nodes_scenarios_times}
\end{figure}
\noindent

The database contains information on technical characteristics of energy system models, including the number of scenarios analyzed, socio-economic indicators such as population figures for specific cities, and techno-economic parameters of individual technologies. The latter category constitutes the majority of the available data.

The number of scenarios considered in energy system analyses provides an indication of the diversity of transformation pathways explored across different regions of the world. The relatively low number of scenarios reported in many studies – typically up to eight (\hyperref[fig:number_nodes_scenarios_times]{Fig. 3c}) – suggests that researchers often focused on a limited set of well-defined futures. More recently, however, growing uncertainties have led to a substantial expansion of scenario analyses. Some studies now evaluate more than 10,000 scenarios, examining cost-optimal pathways, material dependencies, and other system characteristics~\cite{frey2025,müller2026probabilisticidentificationtechnologytipping}. In several regions, particularly parts of Africa, comparatively little attention has so far been devoted to exploring a broad range of scenarios (\hyperref[fig:number_nodes_scenarios_times]{Fig. 3c}). In contrast, the regional close-ups reveal that numerous local energy system scenarios have already been investigated in countries such as the United States, Iran, and China.

The number of scenarios that can be evaluated is closely linked to the complexity of the underlying energy system models. Assessing thousands of scenarios relies on model simplifications, parallel large-scale computing, and supercomputers. Two key dimensions of model complexity are spatial resolution and temporal resolution~\cite{behrens2024reviewing}. 

Spatial granularity, determined by the mean number of nodes, has increased 79\% over the last decade (\hyperref[fig:number_nodes_scenarios_times]{Fig. 3a}), as for example, studies representing Europe as a network of interconnected countries~\cite{hofmann2025h2} have superseded ones that represented it as a single aggregated region~\cite{zeyen2023endogenous}. 
Temporal resolution, defined by the mean number of time steps, influences how accurately dynamic interactions – such as the operation of short-term energy storage systems – can be captured. Our database indicates that the number of modeled yearly time steps increased during the early 2010s (\hyperref[fig:number_nodes_scenarios_times]{Fig. 3b}), reflecting a growing awareness that a limited temporal resolution fails to adequately capture system intermittency, leading to an overestimation of the competitiveness of renewable and baseload technologies and an underestimation of the value of flexibility options~\cite{pfenninger2014energy}. As a result, temporal modeling practices and time step conventions became established in widely used frameworks such as TIMES or OSEMOSYS. 
Advances in time-series aggregation approaches for selecting representative time steps~\cite{KOTZUR2018123} have reduced the need for further increases in temporal resolution, resulting in a plateau in the number of modeled time steps around 2020. Since then, methodological attention appears to have shifted towards improving spatial representation. Continued advances in complexity-reduction techniques~\cite{kotzur2021modeler} and high-performance computing~\cite{frey2025} are therefore enabling researchers to address increasingly sophisticated questions related to energy system transformation.


\section*{Quantitative insights into technology development}\label{Sec-TechnologyInsights}

\begin{figure}[p]
    \centering
    \includegraphics[width=0.96\textwidth]{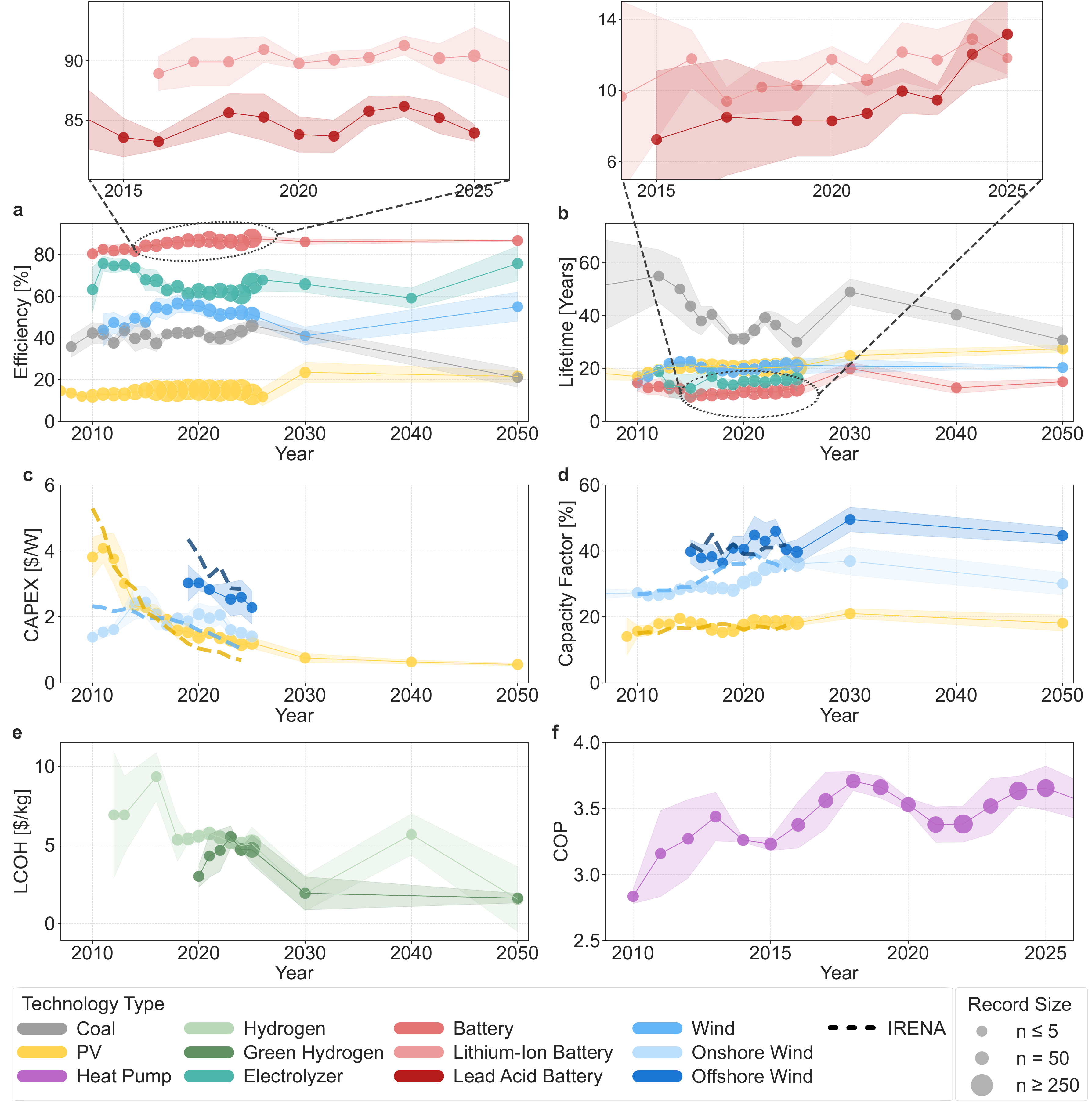}

    \caption{\textbf{Temporal development of six key technological properties evaluated from 5,903 studies.} \textbf{a} and \textbf{b} display the efficiency and lifetime of six key technologies, distinguishing between the two battery subtypes lithium-ion and lead-acid. We excluded all PV efficiency values exceeding the Shockley-Queisser limit ($\sim 33\%$), as such cases represent total system efficiencies or associated with non-PV components of the systems (e.g. solar thermal). \textbf{c} and \textbf{d} show the capital expenditure and capacity factor of PV, onshore and offshore wind systems, compared with reference values reported by the International Renewable Energy Agency (IRENA) Renewable Power Generation Costs Report in 2024~\cite{irena2025renewablecosts}. For onshore wind, 10\% of capex and 12\% of capacity factor of data points are explicitly classified as onshore, while the remaining entries are labelled more generally as wind. \textbf{e} shows the levelized cost of hydrogen and green hydrogen, where green hydrogen (referring to production from renewable electricity). \textbf{f} shows the Coefficient of Performance (COP) of heat pumps covering all types of systems. Values are shown as a centered 3-year moving average over the temporal scope extracted by Quinex or, where unavailable, the publication year. Shaded areas represent the standard error of the centered window, and point size reflects the number of observations per year. To reduce the influence of outliers and sparsely populated years, we applied Tukey's outlier detection method using a multiplier of 3 times the interquartile range (IQR)~\cite{tukey1977exploratory} and retained only years with at least three observations from three distinct publications.}
    \label{fig:tech-properties}
\end{figure}

The database reveals clear structural differences across technologies regarding their assumed efficiency and lifetime (Fig.~\hyperref[fig:tech-properties]{4a,b}). Battery systems are modeled with comparatively short lifetimes of 12 years but high efficiencies of 85\,\%, whereas photovoltaic and wind technologies show substantially longer operational lifetimes above 19 years at lower efficiency levels around 15\,\% and 50\,\%. Coal plant lifetimes show a larger range of values over time, from approximately 30 to 55 years, which likely reflects the small sample size and the presence of several outliers. Some of these outliers~\cite{wang2017malfunction, karatayev2016review} do not refer to the lifetime of coal power plants, but rather to the lifetime of national coal reserves, resulting in substantially higher lifetime values.
These cases highlight the value of the rich metadata provided by our database, as DOI references and surrounding textual context enable the detection and correct interpretation of anomalous values. The database further enables differentiation for example between battery subtypes, with lithium-ion batteries consistently showing higher efficiencies than lead-acid systems and similar lifetime ranges.

To assess the validity of the aggregated extracted values, Figures~\hyperref[fig:tech-properties]{4c} and~\hyperref[fig:tech-properties]{4d} provide a comparison to benchmark data from the International Renewable Energy Agency~(IRENA)~\cite{irena2025renewablecosts}. These show that the extracted data are broadly consistent for capacity factors and confirms the known bias in academic estimates for capital expenditure~(CAPEX).

The inflation-adjusted CAPEX used in academic papers exhibits a pronounced downward trend over time for PV and offshore wind systems, decreasing by $8.8\%/yr$ and $3.8\%/yr$, respectively. While these trends are consistent with the direction of change reported by IRENA, the corresponding declines in the IRENA benchmark are steeper at $13.6\%/yr$ for PV and $8.0\%/yr$ for offshore wind (Fig.~\hyperref[fig:tech-properties]{4c}). In contrast, the CAPEX of onshore wind shows substantial fluctuations and no consistent long-term trend, increasing by only 0.5\%/yr on average, compared to 5.1\%/yr reduction reported by IRENA. Our database reveals systematic differences between assumptions reported in literature and realised technology costs, mirroring the discrepancies previously identified in the energy systems literature~\cite{creutzig2017underestimated, MALHOTRA20202259}. Despite these differences in temporal trends, agreement in absolute values remains strong, with mean absolute errors of 0.4, 0.6 and 0.4 \$/W for onshore wind, offshore wind and PV, respectively.

Similar patterns are observed for renewable capacity factors (Fig.~\hyperref[fig:tech-properties]{4d}). The extracted data align with IRENA’s global average values for PV and offshore wind, while larger discrepancies are apparent for onshore wind between 2015 and 2024. IRENA’s benchmark reports the global weighted-average capacity factor of newly commissioned projects each year, whereas energy system studies typically use values representing the entire installed fleet. Combined with the time delay required for academic publishing, the extracted academic data could therefore be expected to follow the lagged average of the IRENA data. This is observed with a 4-year delay, with the increase from approximately 29\% to 36\% capacity factor occurring from 2015 to 2019 in IRENA’s data, and from 2019 to 2024 in the academic literature. Additional variation arises from scenario-specific modeling assumptions. For instance, some studies model Germany with 100\% renewable power but extensive curtailment, or with exceptionally low wind conditions, yielding capacity factors of 2.0\%~\cite{kies2016curtailment} or 2.3\%~\cite{henckes2018benefit}. Removing such scenarios manually by setting a lower bound of 15\% yields a closer approximation to the IRENA values (Supplementary Fig. S2).

The prospect of hydrogen becoming an important component of future energy systems depends strongly on the levelized cost of hydrogen (LCOH), which remains highly uncertain due to assumptions regarding electricity prices, utilization and capital costs~\cite{chatenet2022water}. The substantial variation in reported LCOH values (Fig.~\hyperref[fig:tech-properties]{4e}) reflects this dependence on underlying techno-economic assumptions.

The economic and environmental performance of heat pumps similarly depends on their coefficient of performance (COP), which varies considerably across operating conditions but is often modeled by a single generic value.
This tendency is reflected in our database, where approximately 12\% of studies assume a COP of 3 and a further 7\% a COP of 4, suggesting a widespread use of round-number assumptions despite the strong context dependence of the metric. Reported COP values have generally increased over the 15 year period, although substantial variation across studies remains (Fig.~\hyperref[fig:tech-properties]{4f}).

\begin{figure}[ht!]
    \centering
    \includegraphics[width=1\linewidth]{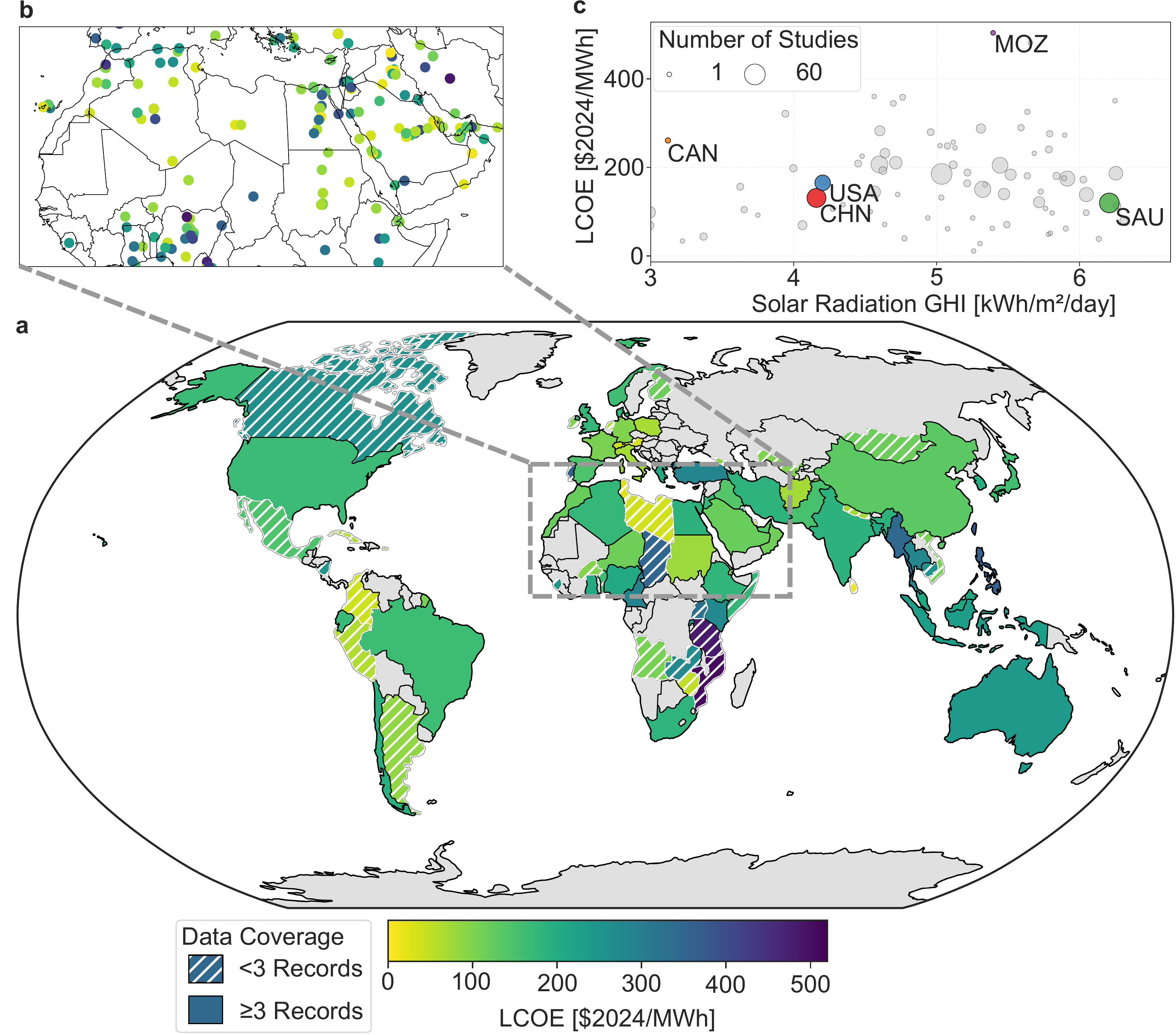}
    \caption{\textbf{Geographical distribution of levelized costs of PV systems drawn from 418 studies.} \textbf{a} shows the country level average LCOE on a world map, while \textbf{b} highlights the precise location of PV studies in North Africa and Southwest Asia. Countries with fewer than three records are shown with hatching. The included PV systems encompass studies with different scopes and publication years. \textbf{c} Relationship between country level LCOE and average daily global horizontal irradiation (GHI), with bubble sizes indicating the number of records per country. The geographic positions are derived from the spatial context extracted by Quinex, and LCOE values are inflation-adjusted to 2024~\$/MWh. LCOE values above 500~\$/MWh are excluded as outliers. To further reduce the influence of outliers, we applied Tukey's outlier detection method using a multiplier of 3 times the interquartile range (IQR)~\cite{tukey1977exploratory}. CAN, CHN, MOZ, SAU, USA correspond to Canada, China, Mozambique, Saudi Arabia, and United States of America, respectively.}
    \label{fig:lcoe-world}
\end{figure}
\noindent

Beyond temporal trends, the spatial metadata in the database enable the analysis of regional patterns in technological characteristics and the identification of underrepresented regions. To illustrate this use case, Fig.~\hyperref[fig:lcoe-world]{5a} presents the country level arithmetic mean levelized cost of electricity (LCOE) of PV systems, derived from the spatial context associated with each data point. In addition, the availability of latitude and longitude information allows for more spatially resolved analyses, as highlighted for North Africa and Southwest Asia (Fig.~\hyperref[fig:lcoe-world]{5b}).

Regional differences in LCOEs are apparent, with Mozambique and Tanzania reporting comparatively high values (Fig.~\hyperref[fig:lcoe-world]{5a}).
These countries' high LCOE values reflect the technological and geographic context of studies conducted in them~\cite{zebra2023scaling, ayeng2019model}. These examine solar home systems and PV mini-grids, which are inherently more expensive per MWh than utility-scale solar farms. This demonstrates the importance of the database enabling context-aware analysis beyond aggregated averages, as this allows researchers to rapidly distinguish between study context and regional conditions when interpreting variations in technology characteristics.

The more fine-grained locations of the PV systems in (Fig.~\hyperref[fig:lcoe-world]{5b}) reveal links to major corridors and nearby urban centres. For example in Saudi Arabia, the observations are aligned along the highway connecting Riyadh, Jeddah and Dammam. In Egypt, a similar pattern is visible, with a concentration of points following the road from Cairo towards the Saharan dessert. In Nigeria, the observations cluster around Kano and Abuja in the dry climate north.

When relating a country's average LCOE to its average solar irradiance, no clear relationship can be observed (Fig.~\hyperref[fig:lcoe-world]{5c}). Countries such as Saudi Arabia (SAU) combine high average solar irradiance with strong research coverage, while Canada (CAN) exhibits both low solar radiation values and limited representation in the literature. In contrast, the United States (USA) and China (CHN) account for a large share of studies despite only moderate average solar radiation values. This likely reflects factors beyond resource availability, including large domestic markets, research capacity, manufacturing bases and individual regions within countries with favourable conditions for PV deployment.

At the same time, countries such as Argentina and Peru, with high average solar irradiance values of 4.9 and 5.1 kWh/m²/day, remain underrepresented in the literature. Similarly, several African countries, including Angola (GHI 5.8) and Mali (GHI 6.0), are underrepresented in the literature. This may reflect broader institutional bias, where research output is concentrated in countries with stronger academic infrastructure and research funding.

\section*{Future expansion and use of the database}\label{Sec-Future}

This article presents our initial energy systems database. By employing a transparent, reproducible, and automated workflow for data expansion, the database can be continuously updated without prohibitive manual effort. Future work will focus on extending the database through more targeted extraction of data from literature on specific technologies. 

Because Quinex is not restricted to a predefined list of technologies or parameters, the database can support exploratory analyses beyond the specific examples presented here. However, as the extraction process does not have a specific, predefined scope, many use cases require filtering the data and grouping it according to common entities, properties, and qualifiers. Future work will facilitate this by automatically linking concepts to ontology classes, for example, of the Open Energy Ontology~\cite{booshehri2021introducing}, which provides a standardised vocabulary for energy systems analysis.

In this work, we use the term database to describe the provision of the extracted data through downloadable data files and an interactive \href{http://dashboard.quinex.org/}{dashboard}. Future work will focus on integrating these resources into a dedicated database infrastructure that further strengthens adherence to widely accepted best practices in research data management. Beyond the automated updates already provided, such a database would provide persistent provenance information and enable researchers to supplement it with their own data, thereby facilitating the exchange of quantitative information across institutions. These developments would further improve the findability, accessibility, interoperability, and reusability of the resulting quantitative information, building upon the FAIR principles~\cite{wilkinson2016fair} that already guide the current implementation.

Large-scale automated extraction does not remove the need for expert judgement. It removes the burden of manual data collection, but human-in-the-loop workflows are necessary for interpretation, filtering on context, and validation. We find that aggregated trends are robust to occasional misclassification of data or the influence of low-quality source material; however, individual records should be assessed using the accompanying metadata, source text and provenance information. This trade-off between scalability and manual curation distinguishes the present database from more narrowly scoped expert-curated resources~\cite{jacobsson2022open, hatton2024global}. The authors of this study recently won the Metascience Novelty Indicators Challenge~\cite{steyn2025novelty}, and we are further developing the approach presented there to augment the metadata associated with each data point in our database with a novelty metric. This is intended to facilitate the identification and selection of suitable data in future applications.

The database infrastructure is designed to be interoperable with energy system modeling workflows. Structured, machine-readable records with provenance metadata could allow automated integration into energy systems modeling frameworks such as ETHOS.FINE~\cite{Klütz2025} and PyPSA~\cite{brown2017pypsa}, integrated assessment models such as GCAM~\cite{calvin2019gcam} or MESSAGE~\cite{huppmann2019messageix}, or with metadata frameworks~\cite{kuckertz2024datadesc, kuckertz2026framework}. This could enable the flexible use of data in energy system models, either through targeted parameter selection or as a basis for large-scale parameter variation studies. Given that many energy system studies have relied on overly pessimistic parameter assumptions for emerging technologies~\cite{ghadim2025we}, the database could help to address this bias in the future.
\section*{Conclusion}\label{Sec-Conclusion}

In this study, we created an open-access database for the energy system domain, compiled from around 76,000 open- and closed-access publications spanning from January 2010 to November 2025. The database encompasses over 3.2 million structured quantitative data points, covering a wide range of entities. Each data point includes metadata, for example its original text passage within the article and, where available, the spatial and temporal context. We highlighted the multiple levels of granularity captured by the database and demonstrated the analytical capabilities of the available metadata. 
The comparison with the external benchmark IRENA further suggests that the extracted values capture meaningful technology trends while also exposing systematic differences in the assumptions used by academics for modeling energy systems. By aggregating quantitative information from the underlying publications, the database substantially reduces the effort required for literature review and data extraction. This enables researchers to focus more on data validation, analysis, modeling and the identification of research gaps.

\section*{Methods}\label{Sec-Methods}

First, we introduce the Quinex framework, which was used to extract quantitative evidence from literature. Subsequently, we describe the workflow used to compile the literature corpus and processing steps applied to prepare the extracted data for the analyses presented in this work.

\subsection*{Information extraction tool}
Quinex~\cite{gopfert2026quinex} is a Python library that uses fine-tuned, open, small language models with 124M and 783M parameters to extract quantitative information from text with high accuracy, demonstrated by F1 scores of over 98\% for quantity identification across diverse domains, as well as 87\% and 82\% for associating the measured properties and entities, respectively.

Quinex first identifies quantity spans (e.g., `9 MW', `three kg', etc.) in the text using an encoder-only transformer model. These spans are then normalized using a rule-based quantity parser, with units mapped to the Quantities, Units, Dimensions and Data Types (QUDT) ontology~\cite{qudt2025}. The context surrounding each quantity span is retained where permitted by copyright restrictions, making it available for 11\% of the quantitative spans, namely those extracted from open-access publications. Quinex identifies associated qualifiers, including the entity and property linked to each quantity, using an encoder-decoder transformer model in a question answering setup. Quinex further detects whether a record has temporal or spatial context and whether it refers to a cited result, while also providing corresponding citation metadata when available. Each extracted record therefore consists of an entity, a property and a quantity, together with optional metadata that can be traced back to the original source document.

Inflation adjustment and currency conversion were performed using CuCoPy~\cite{cucopy}, which retrieves historical price indices from the IMF database~\cite{imf_ifs}. This ensured consistent comparison of techno-economic values across the database.

\subsection*{Workflow for extraction of quantitative evidence}\label{Sec-Extraction}

Between May and November 2025, relevant literature was retrieved from Scopus using the query (below) proposed in a previous bibliometric study on energy systems analysis~\cite{DOMINKOVIC2022111749}, resulting in \textbf{70,034} documents. The equivalent query was implemented in Web of Science (WOS) syntax to capture additional relevant records. The total number of unique publications resulting from this procedure was \textbf{76,306}.

\begin{lstlisting}[label={lst:scopus-query}]
TITLE-ABS-KEY ( ( "energy system*" ) AND ( simul* OR model* OR "plan" OR "plans" OR "planning" OR optimi* OR analy* OR assess* OR evaluat* ) ) AND PUBYEAR > 2009  AND ( LIMIT-TO ( LANGUAGE , "English" ) )
\end{lstlisting}
\noindent
The query targets documents containing the term \textit{energy system} and keywords related to energy systems analysis activities, such as \textit{modeling}, in the title, abstract, or keywords. The query is not restricted to specific document types and includes articles, conference papers and book chapters, among others. However, the query is limited to English-language publications because Quinex currently does not support other languages. 

Subsequently, publications from Elsevier were retrieved directly via the Elsevier API, whereas the remaining publications were downloaded using full-text links obtained through the OpenAlex API. Overall, this approach yielded full-texts for \textbf{31,636} publications. For the remaining documents, the abstract provided by Scopus and WOS was utilised as a fallback, which was successful for \textbf{44,418} documents. For the remaining \textbf{252} documents, neither the full text nor the abstract could be retrieved. The overall collection and processing workflow is summarized in \hyperref[fig:analysis-process]{Fig.~1}.

In the second stage, the full texts and abstracts were processed using Quinex to extract all quantitative statements, resulting in \textbf{3,244,672} data points. While in \textbf{18,531} abstracts no quantitative information was identified, all of the full text included such information. Each extracted record consists of an entity, a property and a quantity. Where available, additional metadata were captured by Quinex, including spatial and temporal context.

As Quinex operates in a domain-agnostic fashion, the raw extraction also includes quantitative information that is not directly related to technology characteristics or techno-economic parameters, such as demographic data (e.g. the population of Beijing). To enable structured analysis and visualisation, the extracted information was mapped to predefined technologies and properties. The technologies and properties selected as examples were chosen based on their frequency of occurrence in the database, as well as their relevance and importance for energy system analysis.

Furthermore, Quinex does not normalise entities and properties, such as 'CAPEX' and 'Capital Cost', into a canonical form. Relevant entities were therefore identified and grouped using regular expressions (regex). For instance, the main category battery was captured with the pattern \texttt{(batter(y|ies))}, whereas the subtype lithium-ion was captured with the pattern \texttt{((lithium.*?ion|li.*?ion|liion).*?(batter(y|ies)))}. A similar approach was applied to other domains, such as wind systems, where distinctions between onshore and offshore wind were employed.

Each entity was compared against all regex patterns and matches were assigned accordingly. If multiple matches occurred within the same main category (e.g. lithium-ion and lead-acid within batteries), the entry was assigned to the general category (battery). However, if matches belonged to different main categories (e.g. onshore wind and lithium-ion battery), no match was assigned to avoid ambiguous classifications. The same procedure was applied to selected properties (e.g. lifetime), thereby enabling consistent aggregation across technologies and characteristics. An overview of selected entities and properties assigned to predefined categories is provided in the Supplementary Table. S1. 

\subsection*{Data and code availability}\label{Sec-Availibility}
An interactive \href{http://dashboard.quinex.org/}{dashboard} is available that allows users to explore the database. Users can download filtered subsets of the raw data or generate visualisations illustrating the temporal evolution and spatial distribution of the selected properties.

The complete raw database is available at \url{https://doi.org/10.26165/JUELICH-DATA/XRI4OH} and all code used for data collection and analysis is publicly accessible at: \url{https://github.com/FZJ-IEK3-VSA/quinex-energy-system-studies}. For publications not licensed under CC BY, the MIT License or the public domain, individual words within text snippets have been masked using a placeholder, and the corresponding Quote field is not provided.


\subsection*{Acknowledgements}\label{Sec-Acknowledgements}

\textbf{Funding}\\
This work was supported by the Helmholtz Association in the context of the Innovation Pool project "Data for Technology Assessment (DaTA)" and by the European Union (ERC, MATERIALIZE, 101076649). Views and opinions expressed are however those of the authors only and do not necessarily reflect those of European Union or the European Research Council Executive Agency. Neither the European Union nor the granting authority can be held responsible for them. This work was also supported by the Helmholtz Association as part of the program “Energy System Design”.
\bigskip
\newline
\textbf{CrediT Author Contributions} \\
Conceptualization: MG, JG, PK, JW \\
Methodology: MG, JG \\
Data curation: MG \\
Formal analysis: MG, JW, IS\\
Funding acquisition: PK, JW, JL\\
Supervision: PK, NH, HH, JW, JL\\
Writing - original draft: MG, JW \\
Writing - review \& editing: MG, JG, PK, NH, HH, IS, JW
\bigskip
\newline
\textbf{AI Declaration}\\
During the preparation of this work the author(s) used ChatGPT and DeepL Write exclusively for language editing and readability improvement. After using this tool, the author(s) reviewed and edited the content as needed and take(s) full responsibility for the content of the publication.
\bigskip
\newline
\textbf{Competing Interests}\\
The authors declare no competing interests. 
\bigskip
\newline
\textbf{Correspondence}\\
Correspondence and requests for materials should be addressed to MG. 

\bibliography{bibliography}

\end{document}


\begin{center}
{\Large \textbf{Supplementary Information}}\\[0.5em]
for\\[0.5em]
{\large Automated Extraction of Techno-Economic Data from 76,000 Energy System Studies}
\end{center}

\renewcommand{\thefigure}{S\arabic{figure}}
\renewcommand{\thetable}{S\arabic{table}}

\newcolumntype{Y}{>{\centering\arraybackslash}X}

\begin{table}[ht]
  \centering
  \setlength{\tabcolsep}{5pt}
  \renewcommand{\arraystretch}{1.4}

  \rowcolors{0}{white}{gray!20}

  \begin{tabularx}{\linewidth}{m{3cm} Y Y Y Y Y}
    \cmidrule(){1-6}
    \rowcolor{white}
    \textbf{Category} & \textbf{Text Phrase 1} & \textbf{Text Phrase 2} & \textbf{Text Phrase 3} & \textbf{Text Phrase 4} & \textbf{$\cdots$} \\
    \cmidrule(){1-6}

    PV
    & PV system & PV & solar PV & PV panels & $\cdots$ \\

    Offshore Wind
    & offshore wind & offshore wind power & offshore wind farms & offshore wind farm & $\cdots$\\

    Lithium-Ion Battery
    & lithium-ion batteries & Li-ion batteries & Li-ion battery & lithium-ion battery & $\cdots$  \\

    Levelized Cost
    & LCOE & COE & LCOH & cost of energy & $\cdots$ \\

    Capacity Factor
    & capacity factor & capacity factors & CF & average capacity factor & $\cdots$  \\

    \cmidrule(){1-6}
  \end{tabularx}

  \caption{\textbf{Examples of extracted text phrases and their mapped categories.} 
  The figure shows the five most frequent extracted entities or properties within each category. Categories are presented independently and are not aggregated across entities or properties. Each category contains additional extracted mentions beyond the examples shown.}
  \label{tab:quinex-table}
\end{table}

\begin{figure}[H]
    \centering
    \includegraphics[width=1\linewidth]{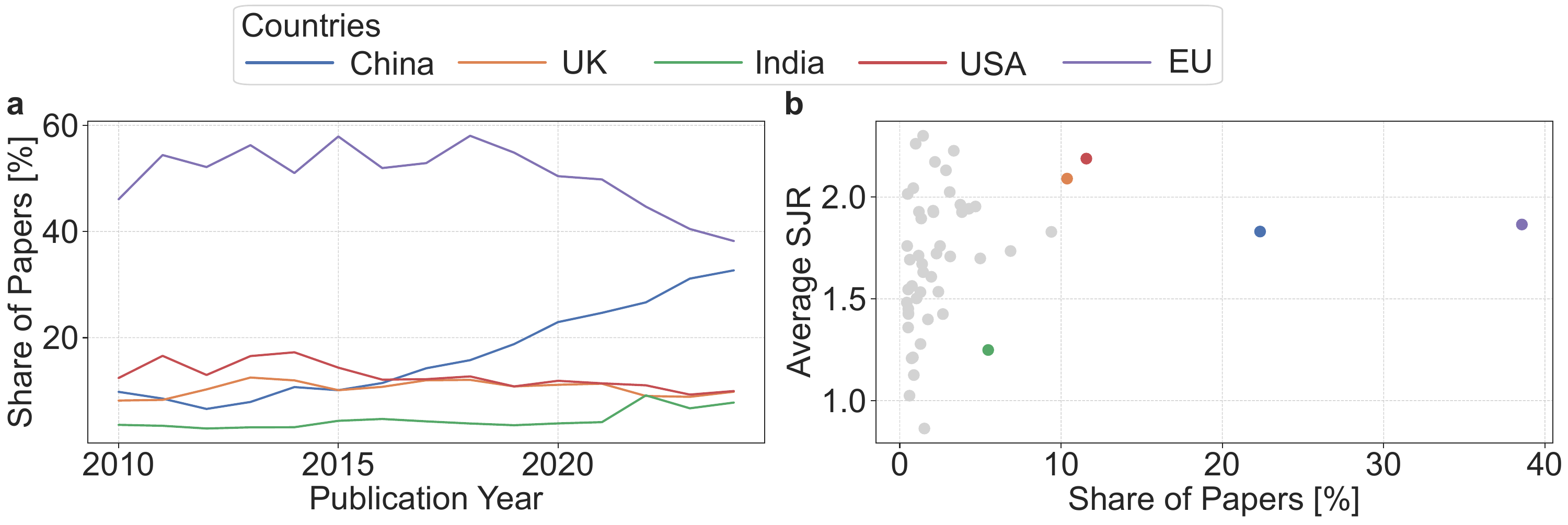}
    
    \caption{\textbf{Geographical origin of the analyzed literature corpus.} The affiliation country information was retrieved via the OpenAlex API~\cite{priem2022openalex}. Publications with multiple country affiliations were fully assigned to each contributing country and publications from 2025
    were excluded because author affiliation information was unavailable for most countries. The country information was missing for around 6\% and SJR scores for around 4\% of publications. \textbf{a} Evolution of the share of full-text publications attributed to the most frequently represented countries. \textbf{b} Relationship between national publication share and average journal impact, measured by SCImago Journal Rank (SJR). CHN, UK, IND, USA, and EU refer to China, United Kingdom, India, United States, and European Union countries, respectively.}
    \label{fig:appendix-country-share}
\end{figure}

\begin{figure}[H]
    \centering
    \includegraphics[width=1\linewidth]{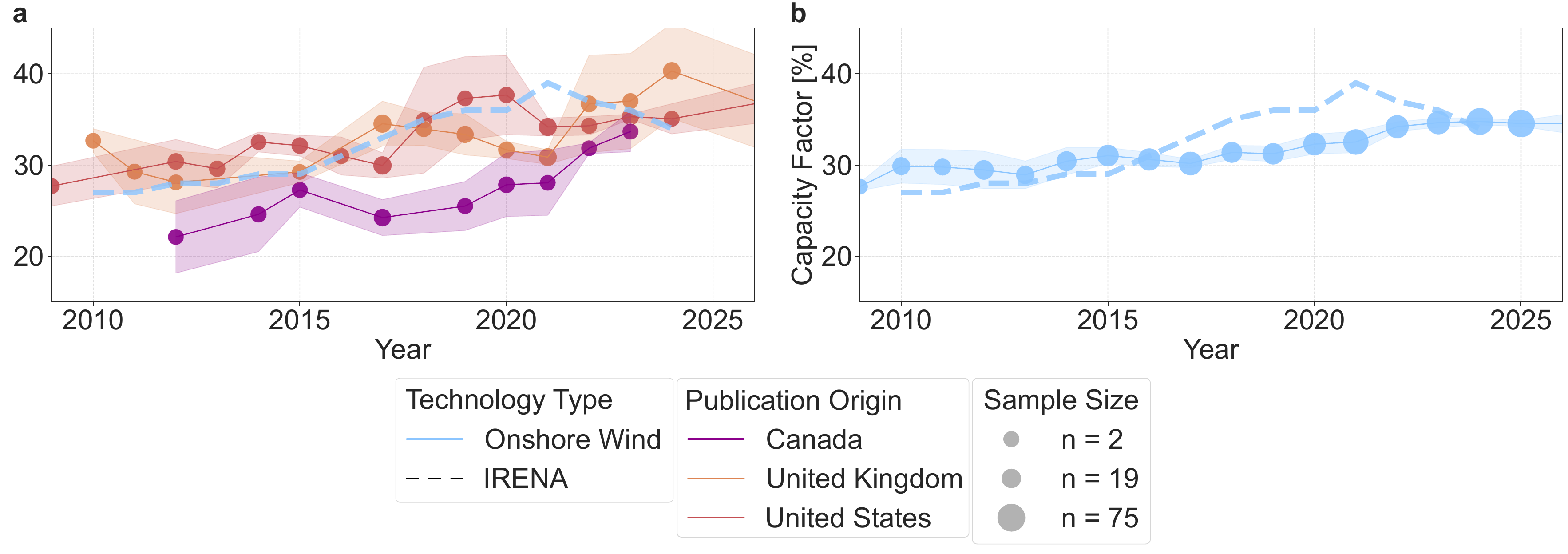}
    
    \caption{\textbf{Detailed temporal evolution of the capacity factor of onshore wind systems.} Values are shown as a centered 3-year moving average over the temporal scope extracted by Quinex or, where unavailable, the publication year. Shaded areas represent the standard error of the centered window, and point size reflects the number of observations per year. To reduce the influence of outliers and sparsely populated years, we applied an outlier filter based on the 20th and 80th percentiles, removing observations outside an extended range (multiplier 1.5) and retaining only years with at least three observations from three distinct publications. \textbf{a} shows the capacity factor reported in publications with the author affiliation originating from Canada, the United Kingdom and the United States. \textbf{b} illustrates the temporal evolution after excluding unrealistic capacity factor values ($\leq15$).}
    \label{fig:appendix-capfactor-wind}
\end{figure}

\bibliography{bibliography}